# Impact Intensity Estimation of a Quadruped Robot without Using a Force Sensor

Ba-Phuc Huynh    Joonbum Bae

Bio-Robotics and Control Laboratory, Department of Mechanical Engineering
Ulsan National Institute of Science and Technology

**Abstract:** Estimating the impact intensity is one of the significant tasks of the legged robot. Accurate feedback of the impact may support the robot to plan a suitable and efficient trajectory to adapt to unknown complex terrains. Ordinarily, this task is performed by a force sensor in the robot's foot. In this letter, an impact intensity estimation without using a force sensor is proposed. An artificial neural network model is designed to predict the motor torques of the legs in an instantaneous position in the trajectory without utilizing the complex kinematic and dynamic models of motion. An unscented Kalman filter is used during the trajectory to smooth and stabilize the measurement. Based on the difference between the predicted information and the filtered value, the state and intensity of the robot foot's impact with the obstacles are estimated. The simulation and experiment on a quadruped robot are carried to verify the effectiveness of the proposed method[1].

## I. INTRODUCTION

Research into the application of legged robots has become popular because of their ability to move flexibly on complex terrains. Many studies have focused on both kinematic structure and motion control to improve robot locomotion [1]-[6].

In the locomotion of a robot, accurate feedback of the landing information is very critical. Most designs use a force sensor placed on the robot foot to measure contact force between the foot and the unknown surface of the terrain [7]-[17]. However, placing the force sensor on the foot is not convenient when the robot has to move in wet, dirty, or chemical environments. Some studies, such as [18], proposed a contact force estimation method based on the generalized momentum of the robot. This method requires the precise calculation of many kinematic and dynamic expressions, which is overly complicated for a multi-legged robot. Some studies, such as [19], just use the position feedback of the servomotors to detect contact between robot legs and terrain surfaces. This method does not determine the magnitude of the impact, so the landing information is limited.

In this letter, an impact intensity estimation of a legged robot without using a force sensor is proposed. Compared with previous studies, this work contains specific novel contributions.

First, an artificial neural network (ANN) model [20], [21] is built to predict the motor torques of no-load (not yet collided with an obstacle) at any instantaneous moving position in the workspace of the robot leg. The inputs to the model are the rotational angles and angular velocities of the motors. The output of the model is the torques of the motors. The actual torque of the motors is measured at the same time and position. If there is no collision, the actual measured value and the predicted value are approximately the same. When a collision occurs, the actual measured value differs from the predicted value. The larger the difference, the greater the impact intensity. Thus, based on the difference between the predicted information and the actual measured value, the state and intensity of the robot foot's impact with the obstacles (or terrain surface) are estimated. Although this method does not determine the accurate value of the contact force, the determination of the impact intensity provides practical information to make the robot motion planning in a more flexible and efficient manner. The training data and the actual measurement data are both collected on the same specific hardware so high precision is achieved. In addition, this method does not require the calculation of complex kinematic and dynamic equations, so it is easy to apply and has a fast execution speed.

Second, an unscented Kalman filter (UKF) is used during the trajectory to smooth and stabilize the measurement of torques. The Kalman filters [22] have been used in many control systems to estimate unknown variables [23]-[24]. UKF [25] uses multiple sigma points to linearize a nonlinear model for more superior accuracy [26]-[31]. Its stability is proved by choosing some initial conditions of the filter [32]. The difficulty in implementing the UKF into the application is primarily in the prediction step of the model. Ordinarily, it requires solving complex system dynamics equations. In the proposed approach, the prediction result from the ANN model is incorporated into UKF to predict the states. This combination simplifies the calculation but achieves the precision required for use in the trajectory control of a legged robot. The proposed ANN-UKF based impact intensity estimation is carried in simulation to adjust filter parameters. It is then applied to a specific quadruped robot to evaluate the effectiveness of the proposed method.

---

[1] Video available at https://www.vinabot.com/2022/04/impact-intensity-estimation-of-.html

The remainder of this letter is organized as follows. Section II introduces the quadruped robot used in the research and the problem statement. Section III describes the motor torque prediction utilizing the ANN model. Section IV presents the the specific design of UKF and the impact intensity estimator. The simulation and experimental results are discussed in Section V. Finally, Section VI is the conclusion and prospects for future work.

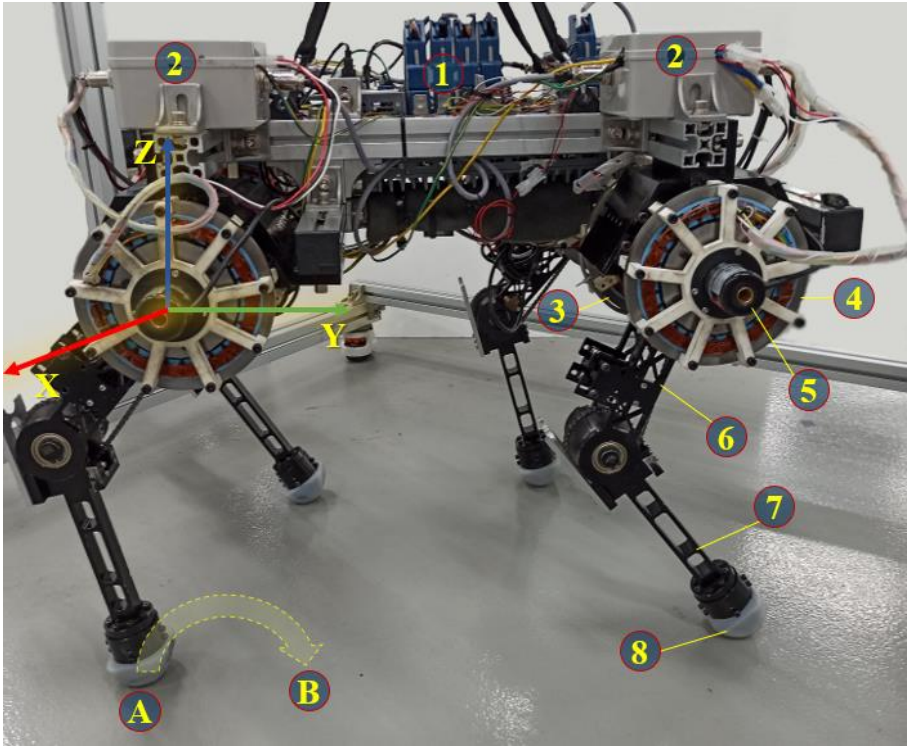

Fig. 1. The quadruped robot used in the proposed method. (1) the controller, (2) the motor drivers, (3) the hip motor, (4) the knee motor, (5) the encoder, (6) the link 1, (7) the link 2, (8) the foot.

## II. PROBLEM STATEMENT

Fig. 1 represents the experimental model of the quadruped robot used in the research. The robot has two-degree-of-freedom legs. Each leg has hip and knee joints, which are driven by the brushless DC motors. The hip and knee motors control the rotation angles of link 1 and link 2, respectively. The motors have incremental encoders to locate the rotation angles. The motors are controlled by the motor driver boards, which can measure the currents through the motors. The measured value of current can be converted to approximate the torque value of the motor based on the manufacturer's specifications. This measurement and conversion are, of course, subject to error. The error is regarded as measurement noise in the UKF process.

When the robot moves, in the swing phase, the footsteps from the start point A to the endpoint B in a curved path. At each moving point in the trajectory, the control state in the joint space of each leg is presented by the vector $p = [\theta_H, \theta_K, \omega_H, \omega_K]^T$, where $\theta_H$, $\theta_K$ are the rotation angles of hip and knee motors, respectively; $\omega_H$, $\omega_K$ are the angular velocities of hip and knee motors, respectively.

In the actual movement of the robot, target position B may not be as desired due to instantaneous high or low terrain surface, or because of obstacles. When a collision occurs, the contact state is specified by the quantity $F$.

The objective of this study is to determine the collision and the value of $F$ without using any force sensor at the foot. The larger the value of $F$, the stronger the collision. In other words, $F$ does not measure the impact force value accurately, it represents whether or not and how strong the impact is. Based on the value of $F$, the trajectory control can determine the terrain surface properties and make more appropriate and effective adjustments. If an accurate value of $F$ is required, a force sensor will be used to calibrate $F$. However, it is unnecessary in case, because we only need to know the impact intensity relatively.

Fig. 2 is the block diagram of the proposed approach. $p_d$ is the desired control state vector provided by the trajectory control. $p_r$ is the real control state vector measured by the controller. $T_m$, $T_p$, and $T_f$ are 2×1 vectors containing the two torque components of the hip and knee motors. $T_p$ is the prediction of the ANN model based on the instantaneous $p_r$. $T_m$ is the real measurement of the controller at $p_r$. The UKF will smooth and stabilize the measurement into the filtered vector $T_f$. The impact intensity estimator will calculate the difference between $T_p$ and $T_f$. The difference value is mapped into the quantity $F$, which will be used as the feedback of the trajectory control.

The following work in this study represents the designs of the ANN model, the UKF, and the impact intensity estimator.

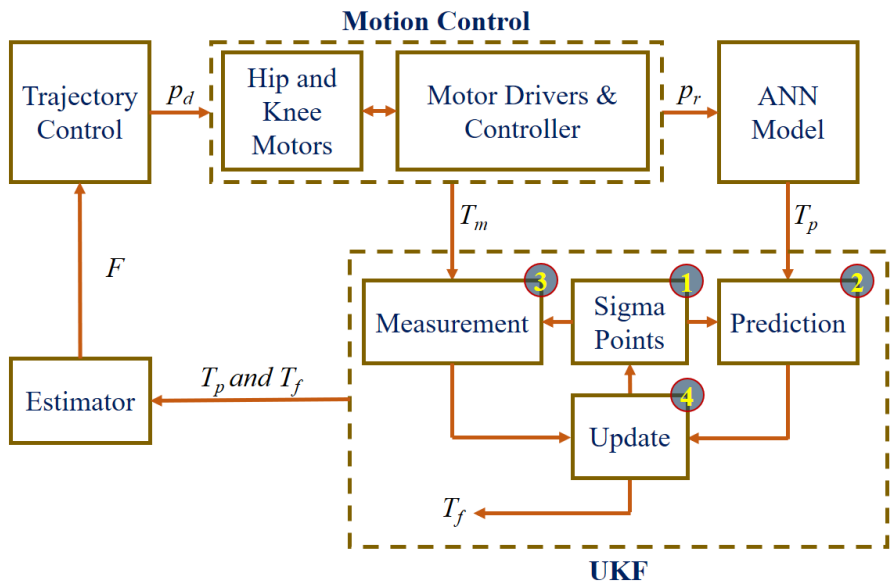

Fig. 2. The block diagram of the control using the impact intensity estimation.

### III. MOTOR TORQUE PREDICTION USING ANN

Fig. 3 is the ANN model used to predict the motor torques. The input of the network is the 4×1 state vector $p$ including rotation angles and angular velocities of the hip and knee motors. The output of the network is the 2×1 vector $T_p$ including two torque components of the hip and knee motors. The network has six hidden layers of 26 neurons and one output layer of 2 neurons with an activation function of tan-sigmoid. The transfer function is tan-sigmoid. Mean squared error (MSE) is used to measure the performance of the network.

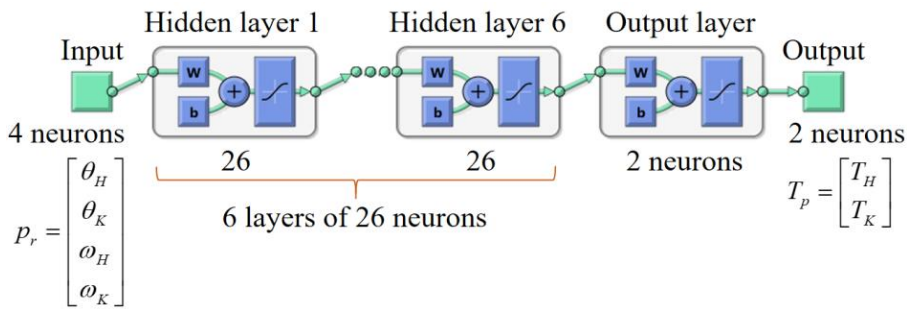

Fig. 3. The ANN model of the motor torque prediction.

The training dataset is acquired by moving the hip and knee motors across locations in the workspace at different angular velocities in no-load states (the robot was hung up so that the legs do not contact the ground). In this study, the training process was done using the *nntool* in MATLAB [33]. The training results in weight and bias factors to calculate an output vector based on a given input vector. It is used to predict the motor torques $T_p$ as presented in Fig. 2.

## IV. UKF AND IMPACT INTENSITY ESTIMATION

### A. The UKF Process

As mentioned in Section I, the state and intensity of the robot foot's impact with the obstacles (or terrain surface) are estimated based on the difference between the predicted information and the actual measured value. It means that based on $T_m$ and $T_p$ in Fig. 2, the estimator can estimate the impact. However, these values include many errors due to noise including the process noise and measurement noise. Therefore, an UKF is designed to smooth and stabilize the measurement before transferring it to the estimator.

The UKF [25] uses multiple sigma points to linearize a nonlinear model for more superior accuracy than the Kalman filters. The primary method is designed for a general problem. Calculation steps and parameters in the method need to be modified to suit a particular problem. The contribution of this paper is to employ the ANN model to predict the state vector in the prediction step to avoid having to solve complex system dynamics expressions. The process of UKF in Fig. 2 includes four main steps, which are applied at each moving point in the trajectory. The process will be stopped when a collision is detected or the trajectory is ended.

**STEP 1: COMPUTATION OF SIGMA POINTS**

The number of sigma points is $m = 2d + 1$, where $d$ is the dimension of the state vector $T_f$. The sigma points are created as follows:

$$\begin{cases} \chi_0 = T_f \\ \chi_{col} = T_f + \left[ \sqrt{(d+\lambda)\mathbf{C_p}} \right]_{col}, \ 1 \leq col \leq d \\ \chi_{col} = T_f - \left[ \sqrt{(d+\lambda)\mathbf{C_p}} \right]_{col-d}, \ d+1 \leq col \leq 2d \\ \lambda = (\alpha^2 - 1)d, \ \alpha > 1 \end{cases} \quad (1)$$

where $\boldsymbol{\chi}$ is a $d \times m$ matrix. $col$ denotes the column of the matrix. Each column of $\boldsymbol{\chi}$ is one sigma point. $\mathbf{C_p}$ is the $d \times d$ covariance matrix. In the initialization (at the start point of the trajectory), $\mathbf{C_p}$ is taken by the process noise matrix $\mathbf{N_p}$. $\lambda$ is the scaling factor. $\alpha$ indicates the difference between the sigma points and the state vector. The Cholesky decomposition can be used to calculate the squared root of the matrices in (1). At the start moving point, $T_f$ is determined by applying the ANN for the start point. At the next moving points, $T_f$ is the previous values calculated in step 4.

The weights are calculated as follows:

$$\begin{cases} \mathbf{W}[r,0] = \dfrac{\lambda}{\lambda + d} \\ \mathbf{W}[r,col] = \dfrac{1}{2(d+\lambda)}, \ 1 \leq col \leq 2d \\ 0 \leq r \leq d-1 \end{cases} \quad (2)$$

where $\mathbf{W}$ are $d \times m$ weight matrice to calculate the mean and the covariance.

**STEP 2: PREDICTION**

The mean value and covariance are predicted as follows:

$$\begin{cases} \hat{f}' = \sum_{col=0}^{2d} \mathbf{W}_{col} \cdot f_{col} \\ \mathbf{C}'_\mathbf{p} = \sum_{col=0}^{2d} \mathbf{W}_{col} \cdot \left[ \left( f_{col} - \hat{f}' \right) \left( f_{col} - \hat{f}' \right)^\mathrm{T} \right] + \mathbf{N}_\mathbf{p} \\ f_{col} = g(\chi_{col}),\ 0 \le col \le 2d \end{cases} \quad (3)$$

where $\mathbf{W}_{col} \in \mathbb{R}^d$ is the $col^{th}$ column of the weight matrix $\mathbf{W}$. The function $g(\cdot)$ is the transition function, which uses the ANN model to predict the motor torque based on $\chi_{col}$ and the instant position $p_r$. $\hat{f}'$ is the mean of prediction. $\mathbf{C}'_\mathbf{p}$ is the predicted covariance matrix. $\mathbf{N}_\mathbf{p}$ is the process noise matrix, which has the size of $d \times d$. $\mathbf{N}_\mathbf{p} = diag(\sigma_p \odot \sigma_p)$, where $\sigma_p \in \mathbb{R}^d$ is the standard deviation vector of the process, which can be estimated based on the accuracy of the ANN model. $\odot$ is the Hadamard product.

The transition function $g(\cdot)$ uses the ANN model to transfer the sigma points to the prediction space as follows:

$$\begin{cases} g(\chi_{col}) = \chi_{col} + diff(T_f, T_p) + v_p \\ T_p = \mathrm{ANN}(p_r) \end{cases} \quad (4)$$

where $\mathrm{ANN}(p_r)$ calculates the predicted torque vector $T_p$ at the instant position $p_r$ by using the ANN model. The *diff* operation calculates the difference between the new predicted torque vector and the previous torque vector. $v_p$ is the additive process noise.

**STEP 3: MEASUREMENT**

The mean value and covariance of the measurement are calculated as follows:

$$\begin{cases} \hat{z} = \sum_{col=0}^{2d} \mathbf{W}_{col} \cdot z_{col} \\ \mathbf{C}_\mathbf{z} = \sum_{col=0}^{2d} \mathbf{W}_{col} \cdot \left[ (z_{col} - \hat{z})(z_{col} - \hat{z})^\mathrm{T} \right] + \mathbf{N}_\mathbf{m} \\ z_{col} = h(\chi_{col}),\ 0 \le col \le 2d \end{cases} \quad (5)$$

where the function $h(\cdot)$ is the transition function to map the sigma points to the measurement space. $\hat{z}$ is the mean of measurement. $\mathbf{C}_\mathbf{z}$ is the covariance matrix in the measurement space. $\mathbf{N}_\mathbf{m}$ is the measurement noise matrix, which has the size of $d \times d$. $\mathbf{N}_\mathbf{m} = diag(\sigma_m \odot \sigma_m)$, where $\sigma_m \in \mathbb{R}^d$ is the standard deviation vector, which can be estimated based on the accuracy of the motor driver and encoder. $\odot$ is the Hadamard product.

The function $h(\cdot)$ is similar to the function $g(\cdot)$ but replaces the additive process noise $v_p$ with the additive measurement noise $v_m$.

**STEP 4: UPDATE STATES**

The cross-covariance matrix is calculated as follow:

$$\mathbf{C}_\mathbf{v} = \sum_{col=0}^{2d} \mathbf{W}_{col} \cdot \left[ (f_{col} - \hat{f}')(z_{col} - \hat{z})^\mathrm{T} \right] \quad (6)$$

The Kalman gain is determined as follow: :

$$\mathbf{K} = \frac{\mathbf{C}_\mathbf{v}}{\mathbf{C}_\mathbf{z}} \quad (7)$$

Finally, the output and covariance matrix are updated as follows:

$$T_f = \hat{f}' + \mathbf{K}(T_m - \hat{z})$$
$$\mathbf{C_p} = \mathbf{C'_p} - \mathbf{K}\mathbf{C}_z\mathbf{K}^T \quad (8)$$

*B. Impact intensity Estimation*

At each moving point of the trajectory, the difference between the filtered torque vector and the predicted torque vector is calculated as follow:

$$T_{diff} = \begin{bmatrix} T_H \\ T_K \end{bmatrix} = \max([0,0]^T, |T_f - T_p| - T_{threshold}) \quad (9)$$

where $T_{threshold} \in \mathbb{R}^d$ is to compensate for the model error. It can be estimated in the experiment based on the accuracy of the ANN model.

The difference of torque is then mapped to the contact force as follow:

$$\vec{F} = \begin{bmatrix} \vec{F}_H \\ \vec{F}_K \end{bmatrix} = \Gamma(T_{diff}) \quad (10)$$

where $\vec{F}_H$, $\vec{F}_K$ are the component forces on the robot leg corresponding to the torques of the hip and knee motors, respectively. $\Gamma$ is the function to calculate the force based on the torques and the lever arms (link 1 and link 2 in Fig. 1).

$\vec{F}$ represents the impact force of the foot of the robot as mentioned in Section II. The collision is detected when any element of $F$ is greater than zero.

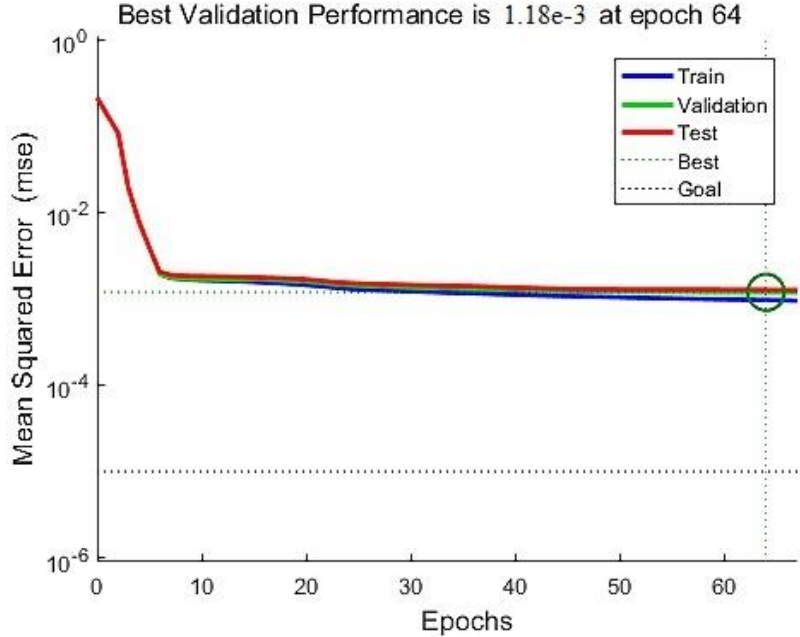

Fig. 4. The performance of ANN training.

## V. SIMULATION AND EXPERIMENTAL RESULT

*A. The ANN training*

The training dataset consists of 9708 samples evenly distributed throughout the robot leg's workspace. The input includes the rotation angle of the hip motor (rad), the rotation angle of the knee motor (rad), the angular velocity of the hip motor (rad/second), and the angular velocity of the knee motor (rad/second). The output includes the currents of hip and knee motors (amp). After the training, the output is mapped to the motor torque based on the specification of the hardware. The goal of performance (MSE) in training is set to 1e-5. Fig. 4 displays the performance of the training. The best validation performance is 1.18e-3 after 64 epochs. This value is appropriate enough for the model because the current of the motors is quite large when the robot works. Fig. 5 presents the regression of the model. The regression value is close to 1, which indicates a proper fit of the model. In fact, in practice, we do not need the very high accuracy of the ANN model because the contact force is estimated based on the difference between the prediction and the filtered measurement. The UKF combines the prediction of the ANN model and the real measurement to smooth and stabilize the measured value, then detect the collision. In the test on the training dataset, the maximum error of the output is 0.1208 (amp). This value is manipulated to estimate the standard deviation of the process noise $\mathbf{N}_p$ in the UKF as presented in Section IV.A.

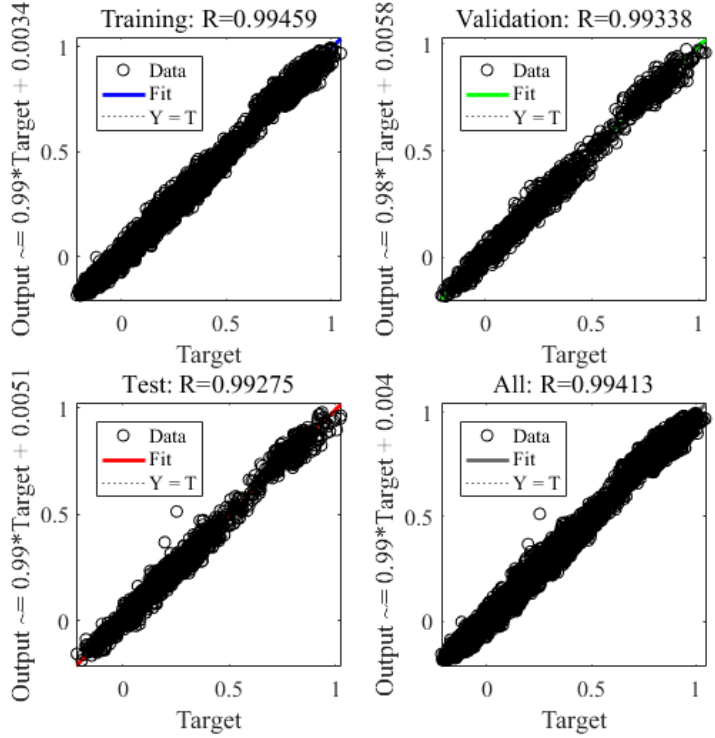

Fig. 5. The regression of ANN training.

*B. Simulation*

The purpose of the simulation is to determine suitable parameters for the UKF. The simulation was performed in MATLAB. Fig. 6 is the trajectory of the foot of the robot in the swing phase. Fig. 7 presents the tracking data of the simulation. The predicted value is calculated by the ANN model based on the position and velocity of hip and knee motors. The predicted value is equally considered as the apparent ground truth of the model. The real measurement is emulated by adding a random noise into the predicted value. The random range is 0 to 10% of the value, which is based on the real hardware. This value is additionally used to estimate the standard deviation of the measurement noise in the UKF. $\alpha$ in (1) is 10 for proper performance. The program launches a collision in the trajectory, where the measurement becomes large. Before the collision, the filtered value of the UKF is close to the prediction value although there is frequent noise in the measurement. When the collision occurs, the filtered value of the UKF is clearly different from the predicted value. This difference allows the program to detect the collision and estimate the contact force as described in Section IV.B.

*C. Experiment*

The quadruped robot is used to evaluate the proposed method as presented in Fig. 1. The main controller is the module NI cRIO-9039. The control program is made in LabWindows. Based on the simulation, the parameters of the UKF are set as in

Table I. The leg of the robot steps from the start point to the endpoint with a similar trajectory as in Fig. 6. The z-coordinate of the endpoint is set to -460 mm, which is lower than the ground to cause a collision between the foot and the ground at the end.

Fig. 8 presents the tracking data of the experiment. Before the collision, the filtered values of the UKF are close to the prediction of the ANN model although the measurement noise is large. When the foot of the robot contacts the ground surface, the measurement becomes larger. The filtered value begins to move away from the predicted value. This difference is enough for the program to recognize the collision and stop the motion. However, in this experiment, the program continued to move the robot's leg for the rest of its trajectory. This causes the robot's leg to hit the ground and raise the robot's body slightly. It accounts for the slight fluctuation of the filtered value after the collision. After that, the filtered value starts to increase in proportion to the increased impact force on the robot foot.

Table I
The UKF parameters

| Parameter | Value | Unit |
|---|---|---|
| $\alpha$ | 10 | |
| $\sigma_p$ | 8.5 | N.cm |
| $\sigma_m$ | 178 | N.cm |
| $v_p$ | [-3.4, 3.4] | N.cm |
| $v_m$ | [-3.4, 3.4] | N.cm |
| $T_{threshold}$ | $(1,1)^T$ | N.cm |

Table II
The summary of the difference between the filtered value and the prediction of the ANN model before the collision

| Data | Hip motor | | Knee motor | |
|---|---|---|---|---|
| | Min | Max | Min | Max |
| Simulation | 0 | 2.2558 | 0 | 2.2558 |
| Experiment | 0.0023 | 5.0253 | 0.0020 | 3.7677 |

Table II summarizes the difference between the filtered value and the prediction before the collision in both simulation and experiment. Based on this data, the value of $T_{threshold}$ in (9) can be determined to estimate the impact intensity.

## VI. CONCLUSION AND FUTURE WORK

In this letter, an impact intensity estimation for legged robots has been presented. The ANN model was developed to predict the motor torques based on the specific position and angular velocity of the motors. The motor torque is calculated by the motor current measurement. This process includes a significant noise. However, the UKF smoothed and stabilized the measurement. Based on the difference between the filtered value of the UKF and the prediction of the ANN model, the impact intensity was estimated. The integration of the ANN model into the UKF helps to detect collisions and estimate the impact intensity easily without using any force sensor on the foot of the robot. Simulation and experiment on a quadruped robot verified the effectiveness of the proposed method.

The proposed method was implemented to provide information on the landing of the robot. In future work, this information will be used to adjust the trajectory of the robot to adapt to a variety of complex and unknown terrain surfaces. Although the experiment was implemented on a legged robot, the proposed method may be utilized for other systems to detect the collision and estimate the contact force without using force sensors.

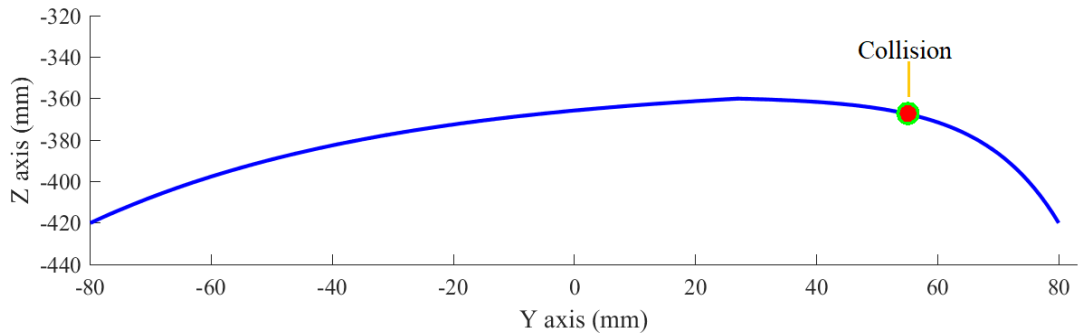

Fig. 6. The trajectory of the foot of the quadruped robot in the swing phase.

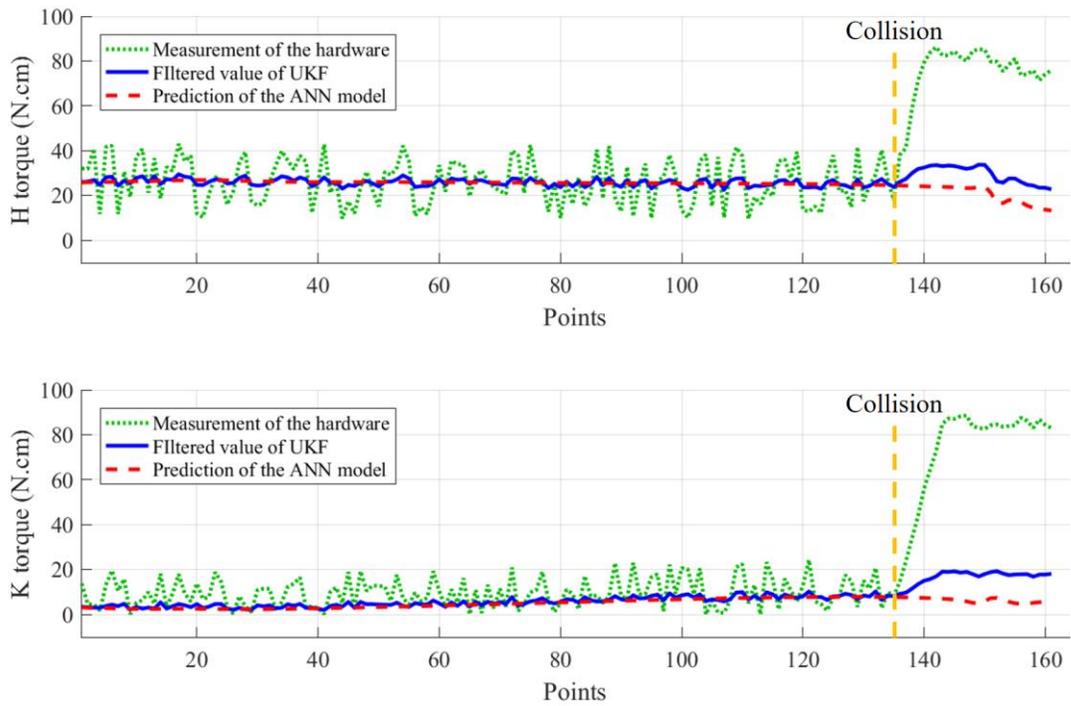

Fig. 7. The tracking data of the hip and knee motors in simulation.

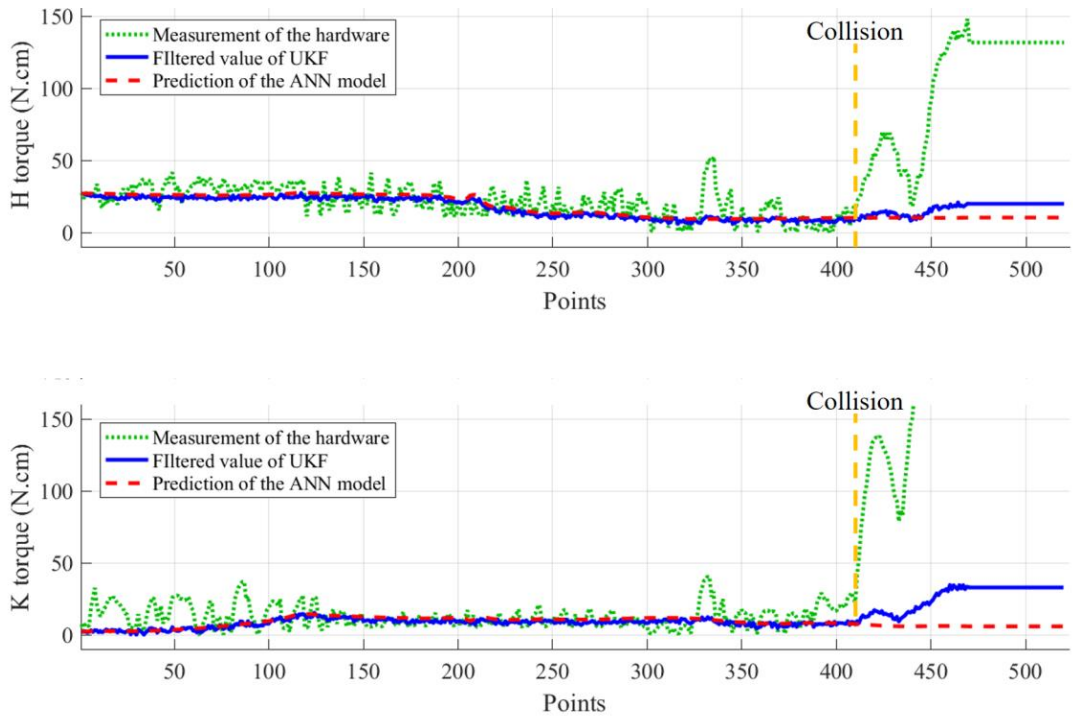

Fig. 8. The tracking data of the hip and knee motors in the experiment.

**REFERENCES**

[1] Pierre-Brice Wieber, Russ Tedrake, Scott Kuindersma, "Modeling and Control of Legged Robots," Springer Handbook of Robotics, Springer International Publishing, pp.1203-1234, 2016.
[2] J. Sun and J. Zhao, "An Adaptive Walking Robot With Reconfigurable Mechanisms Using Shape Morphing Joints," *IEEE Robotics and Automation Letters*, Vol. 4, No. 2, pp. 724-731, Apr. 2019.
[3] S. Aoi, P. Manoonpong, Y. Ambe, F. Matsuno, and F. Wörgötter, ¨"Adaptive control strategies for interlimb coordination in legged robots: A review," *Frontiers Neurorobotics*, vol. 11, Art. 39, Aug. 2017.
[4] J. Chen, F. Gao, C. Huang, and J. Zhao, "Whole-Body Motion Planning for a Six-Legged Robot Walking on Rugged Terrain," *Applied Sciences*, Vol. 9, No. 24, 5284, Dec. 2019.
[5] D. Belter, J. Wietrzykowski, and P. Skrzypczyński, "Employing Natural Terrain Semantics in Motion Planning for a Multi-Legged Robot," *Journal of Intelligent & Robotic Systems*, Vol. 93, pp. 723–743, May 2018.
[6] B. Hammoud, M. Khadiv, and L. Righetti, "Impedance Optimization for Uncertain Contact Interactions Through Risk Sensitive Optimal Control," *IEEE Robotics and Automation Letters*, Vol. 6, No. 3, pp. 4766-4773, Jul. 2021.
[7] H. Xia, X. Zhang, and H. Zhang, "A New Foot Trajectory Planning Method for Legged Robots and Its Application in Hexapod Robots,", *Applied Sciences*, Vol. 11, 9217, pp. 1-19, Oct. 2021.
[8] J. Carius , R. Ranft, V. Koltun, and M. Hutter, "Trajectory Optimization for Legged Robots With Slipping Motions," *IEEE Robotics and Automation Letters*, Vol. 4, No. 3, pp. 3013-3020, Jul. 2019.
[9] S. Dutta, T. K. Maiti, M. M-Mattausch, Y. Ochi, N. Yorino, and H. J. Mattausch, "Analysis of Sensor-Based Real-Time Balancing of Humanoid Robots on Inclined Surfaces," *IEEE ACCESS*, Vol. 8, pp. 212327-212338, Nov. 2020.
[10] J.H Kim, "Multi-Axis Force-Torque Sensors for Measuring Zero-Moment Point in Humanoid Robots: A Review," *IEEE Sensors Journal*, Vol. 20, No. 3, pp 1126-1141, Feb. 2020.
[11] G. Valsecchi, R. Grandia, and M. Hutter, "Quadrupedal Locomotion on Uneven Terrain With Sensorized Feet," *IEEE Robotics and Automation Letters*, Vol. 5, No. 2, pp. 1548-1555, Apr. 2020.
[12] S. Fahmi, M. Focchi, A. Radulescu, G. Fink, V. Barasuol, and C. Semini, "STANCE: Locomotion Adaptation Over Soft Terrain," *IEEE Transactions on Robotics*, Vol. 36, No. 2, pp. 443-457, Apr. 2020.
[13] X. A. Wu, T. M. Huh, A. Sabin, S. A. Suresh, and M. R. Cutkosky, "Tactile Sensing and Terrain-Based Gait Control for Small Legged Robots," *IEEE Transactions on Robotics*, Vol. 36, No. 1, pp. 15-27, Feb 2020.
[14] F. Grimminger, A. Meduri, M. Khadiv, J. Viereck, M. Wüthrich, M. Naveau, V. Berenz, S. Heim, F. Widmaier, T. Flayols, J. Fiene, A. B-Spröwitz, and L. Righetti, "An Open Torque-Controlled Modular Robot Architecture for Legged Locomotion Research," *IEEE Robotics and Automation Letters*, Vol. 5, No. 2, pp. 3650-3657, Apr. 2020.
[15] C. Tiseo, W. Merkt, K. K. Babarahmati, W. Wolfslag, I. Havoutis, S. Vijayakumar, and M. Mistry, "HapFIC: An Adaptive Force/Position Controller for Safe Environment Interaction in Articulated Systems," *IEEE Transactions on Neural Systems and Rehabilitation Engineering*, Vol. 29, pp. 1432-1440, 2021.
[16] Q. Li, Z. Yu, X. Chen, Q. Zhou, W. Zhang, L. Meng, and Q. Huang, "Contact Force/Torque Control Based on Viscoelastic Model for Stable Bipedal Walking on Indefinite Uneven Terrain," *IEEE Transactions on Automation Science and Engineering*, Vol. 16, No. 4, pp. 1627-1639, Oct. 2019.
[17] E. Sihite, P. Dangol, and A. Ramezani, "Unilateral Ground Contact Force Regulations in Thruster-Assisted Legged Locomotion," in *IEEE/ASME International Conference on Advanced Intelligent Mechatronics (AIM)*, May 2021, pp. 389-395.
[18] Z. Cong , H. An, C. WU, L. Lang, Q. Wei, and H. Ma, "Contact Force Estimation Method of Legged-Robot and Its Application in Impedance Control," *IEEE ACCESS*, Vol. 8, pp. 161175-161187, 2020.
[19] P. Čížek, M. Zoula, and J. Faigl, "Design, Construction, and Rough-Terrain Locomotion Control of Novel Hexapod Walking Robot With Four Degrees of Freedom Per Leg," *IEEE ACCESS*, Vol. 9, 17866-17881, Feb. 2021.
[20] J. L. G. Rosa, *Artificial Neural Networks: Models and Applications*. London, United Kingdom: InTechOpen, Oct. 2016.
[21] A. Y. Alanis, N. Arana-Daniel, and C. López-Franco, *Artificial Neural Networks for Engineering Applications*, 1st ed. Academic Press, Feb. 2019.
[22] R. E. Kalman, "A New Approach to Linear Filtering and Prediction Problems," *Transactions of the ASME–Journal of Basic Engineering*, Vol.82, pp. 35-45, 1960.
[23] K. Ansari, "Real-Time Positioning Based on Kalman Filter and Implication of Singular Spectrum Analysis," *IEEE Geoscience and Remote Sensing Letters*, Vol. 18, No. 1, pp. 58-61, Jan. 2021.
[24] D. Sun, X. Li, Z. Cao, J. Yong, D. Zhang, and J. Zhuang, "Acoustic Robust Velocity Measurement Algorithm Based on Variational Bayes Adaptive Kalman Filter," *IEEE Journal of Oceanic Engineering*, Vol. 46, No. 1, pp. 183-194, Jan. 2021.
[25] S.J. Julier and J.K. Uhlmann, "New Extension of the Kalman Filter to Nonlinear Systems," *Proc. SPIE, Signal Processing, Sensor Fusion, and Target Recognition VI*, Vol. 3068, pp. 182-193, Jul. 1997.
[26] J. Zhao and L. Mili, "A Decentralized H-Infinity Unscented Kalman Filter for Dynamic State Estimation Against Uncertainties," *IEEE Transactions on Smart Grid*, Vol. 10, No. 5, pp. 4870-4880, Sep. 2019.

[27] I. Ullah, Y. Shen, X. Su, C. Esposito, and C. Choi, "A Localization Based on Unscented Kalman Filter and Particle Filter Localization Algorithms," *IEEE ACCESS*, Vol. 8, pp. 2233-2246, Jan. 2020.

[28] J. Li, M. Ye, S. Jiao, W. Meng, and X. Xu, "A Novel State Estimation Approach Based on Adaptive Unscented Kalman Filter for Electric Vehicles," *IEEE ACCESS*, Vol. 8, pp. 185629-185637, Oct. 2020.

[29] Z. Xu, S. X. Yang, and S. A. Gadsden, "Enhanced Bioinspired Backstepping Control for a Mobile Robot With Unscented Kalman Filter," *IEEE ACCESS*, Vol. 8, pp. 125899-125908, Jul. 2020.

[30] Y. E. Kim, H. H. Kang, and C. K. Ahn, "Two-Layer Nonlinear FIR Filter and Unscented Kalman Filter Fusion With Application to Mobile Robot Localization," *IEEE ACCESS*, Vol. 8, pp. 87173-87183, May 2020.

[31] R. Yildiz, M. Barut, and E. Zerdali, "A Comprehensive Comparison of Extended and Unscented Kalman Filters for Speed-Sensorless Control Applications of Induction Motors," IEEE Transactions on Industrial Informatics, Vol. 16, No. 10, pp. 6423-6432, Oct. 2020.

[32] L. Li and Y. Xia, "Stochastic Stability of the Unscented Kalman Filter with Intermittent Observations," *Automatica*, Vol. 48, pp. 978-981, 2012.

[33] X. Pang, "Application of Matlab/NNTool in Neural Network System," *Computer Science*, Vol. 4, pp. 125-128, 2004.